\documentclass[10pt,journal,final]{IEEEtran}
\usepackage{amsmath,amsfonts}
\usepackage{array}
\usepackage{algorithmic}
\usepackage[ruled,lined,linesnumbered,ruled]{algorithm2e} 
\usepackage{multirow}
\usepackage{graphicx} 
\usepackage{float} 
\usepackage[caption=false,font=footnotesize,labelfont=rm,textfont=rm]{subfig}
\usepackage{CJKutf8}
\usepackage{textcomp}
\usepackage{stfloats}
\usepackage{url}
\usepackage{verbatim}
\usepackage{graphicx}
\usepackage{cite}
\begin{document}
\begin{CJK}{UTF8}{gbsn}

\title{\textbf{Self-Organizing Recurrent Stochastic Configuration Networks for Nonstationary Data Modelling}}

\author{
  Gang Dang \\
  State Key Laboratory of Synthetical Automation for Process Industries \\
  Northeastern University, Shenyang 110819, China\\
   Dianhui Wang 
  \thanks{\textit{\underline{Corresponding author}}: 
\textbf{dh.wang@deepscn.com}}\\
  State Key Laboratory of Synthetical Automation for Process Industries \\
  Northeastern University, Shenyang 110819, China \\
 Research Center for Stochastic Configuration Machines\\
  China University of Mining and Technology, Xuzhou 221116, China\\
}

\maketitle
\newtheorem{remark}{\bf Remark}       

\begin{abstract}                          
Recurrent stochastic configuration networks (RSCNs) are a class of randomized learner models that have shown promise in modelling nonlinear dynamics. In many fields, however, the data generated by industry systems often exhibits nonstationary characteristics, leading to the built model performing well on the training data but struggling with the newly arriving data. This paper aims at developing a self-organizing version of RSCNs, termed as SORSCNs, to enhance the continuous learning ability of the network for modelling nonstationary data. SORSCNs can autonomously adjust the network parameters and reservoir structure according to the data streams acquired in real-time. The output weights are updated online using the projection algorithm, while the network structure is dynamically adjusted in the light of the recurrent stochastic configuration algorithm and an improved sensitivity analysis. Comprehensive comparisons among the echo state network (ESN), online self-learning stochastic configuration network (OSL-SCN), self-organizing modular ESN (SOMESN), RSCN, and SORSCN are carried out. Experimental results clearly demonstrate that the proposed SORSCNs outperform other models with sound generalization, indicating great potential in modelling nonlinear systems with nonstationary dynamics.
 
\end{abstract}

\vspace{-0.3cm}

\section{Introduction}
\IEEEPARstart{M}{odern} industrial processes often exhibit nonlinear, uncertain, and multi-variable evolutionary behaviours. The sample data generated from such complex systems may be dynamically changing and nonstationary \cite{ref0,ref1,ref2}, posing significant challenges in building an accurate prediction model. Neural networks (NNs) have received considerable attention in modelling nonlinear dynamics due to their powerful approximation performance, high fault tolerance, and adaptive learning capability \cite{ref3,ref4,ref5}. In \cite{ref6}, the convolutional neural networks were utilized to extract nonstationary time series features and fuse them across cascade levels to improve the representation of the process data. Long short-term memory (LSTM) network can capture long-term dependent relationships in signal sequences, focusing on the inherent information between them, which proves to be effective in handling the changing temporal data \cite{ref7}. Additionally, some results on the industrial data analysis using the evolutionary algorithm and self-organizing learning strategy have been reported in \cite{ref9,ref10,ref101,ref102}, which can dynamically adjust the parameters and structure of the network based on real-time data streams. Although these approaches mentioned are effective in modelling nonstationary dynamics, a common problem persists. Most of these models rely on the error back-propagation (BP) algorithm to train NNs, which suffers from slow convergence, structure determination, and the sensitivity of the initial weights and the learning parameters.

In recent years, randomized methods for training NNs have gained considerable interest due to their fast learning speed and computational simplicity \cite{ref103,ref104,ref105}. In \cite{ref11}, a self-organizing version of the random vector functional link network (RVFLN), termed as parsimonious RVFLN, was presented to automatically configure the network structure. While parsimonious RVFLN demonstrated superior performance in data streaming tasks, the approximation ability of the network cannot be guaranteed if the random parameters setting is unreasonable. Wang and Li \cite{ref12} pioneered a novel randomized learner model, termed stochastic configuration networks (SCNs), which innovatively introduced a supervisory mechanism to generate the random parameters, ensuring the universal approximation property. Such a finding is significant to the development of randomized learning theory, and many promising results about modelling complex dynamics have been reported \cite{ref13,ref14,ref15}. In \cite{ref16}, a deep stacked SCN was proposed, which offered a handy implemented framework with universal approximation capability for the continuous learning of nonstationary data streams. Li et al. \cite{ref17} developed the online self-organizing SCNs, which can adaptively adjust the network structure and parameters based on the real-time arriving data. Unfortunately, these methods are built on the assumption of the known system orders. In real industrial applications, the system orders are time-varying, resulting in order uncertainty. Echo state networks (ESNs) provide an alternative methodology for modelling nonlinear systems with unknown dynamical orders \cite{ref19}. They utilize a large-scale sparsely connected reservoir to process the temporal information, and their self-organizing variants have demonstrated great potential in processing the dynamical time series data \cite{ref20,ref21,ref22}. However, ESNs lack a basic understanding on both the structure setting and the random parameters assignment. Recent work in \cite{ref221} motivates us to look into modelling nonstationary dynamics using an advanced randomized learner model, termed as recurrent stochastic configuration networks (RSCNs), which are incrementally built by assigning the random parameters in the light of a supervisory mechanism.

This paper proposes a self-organizing version of RSCNs (SORSCNs) for modelling nonlinear systems with nonstationary data streams. Given a collection of historical data, the initial model is constructed through a supervisory mechanism, and an error interval is determined based on the training result. Then, the built model is utilized to predict newly arriving data streams and obtain the output error $e$. If $e$ is within the preset error interval, it indicates that the network can effectively capture information from the newly data, and the projection algorithm is used to update the readout weights online. If $e$ is larger than the upper bound of the preset error interval, the reservoir structure needs to be dynamically adjusted based on the recurrent stochastic configuration (RSC) algorithm and an improved sensitivity analysis. The main contributions of this work can be summarized as follows.

\begin{itemize}
  \item [1)] 
 A hybrid self-learning framework is presented, where the characteristics of real-time data streams are considered to dynamically adjust the network parameters and structure. By employing a unique growing-pruning-growing scheme, the model's capacity for continuous learning in response to unknown nonstationary dynamics is enhanced.      
  \item [2)]
The universal approximation property and echo state property of the original RSCN are naturally inherited during the construction.
  \item [3)]
The developed method is applied to two real-world industrial data predictive analyses. The results demonstrate the superiority of our proposed SORSCNs in terms of prediction accuracy and structural compactness.
\end{itemize}

The remainder of this paper is organized as follows. Section II briefly reviews the related knowledge of RSCNs. Section III details our proposed SORSCNs. Section IV reports the experimental results. Finally, Section V concludes this paper and provides some future research directions.

\section{A brief review of RSCNs}
This section briefly reviews RSCNs \cite{ref221}, which hold the universal approximation capability for any given nonlinear temporal data. RSCNs start with a small-sized network, which is constructed in the light of a supervisory mechanism. Due to their merits in neural network construction, such as less human intervention, adaptability of random parameters scope setting, fast learning, and handy implementation, RSCNs have demonstrated great potential in modelling nonlinear complex dynamics. Moreover, to accelerate the convergence of training errors and facilitate the implementation of the self-organizing strategy, the block incremental recurrent stochastic configuration network (BRSCN) is exploited in this paper, where each block can be viewed as a subreservoir. 

Given the input $\mathbf{U}\text{=}\left[ \mathbf{u}(1),\ldots ,\mathbf{u}({{n}_{\max }}) \right]$ and desired output $\mathbf{T}=\left[ \mathbf{t}\left( 1 \right),...\mathbf{t}\left( {{n}_{max}} \right) \right]$, assume that we have built a BRSCN model with $j$ subreservoirs, that is,

\begin{scriptsize}
\vspace{-0.2cm}
\begin{equation} \label{eq11}
\begin{array}{*{20}{c}}
\begin{array}{l}
\left[ {\begin{array}{*{20}{c}}
{{{\bf{x}}^{\left( 1 \right)}}(n)}\\
{{{\bf{x}}^{\left( 2 \right)}}(n)}\\
 \vdots \\
{{{\bf{x}}^{\left( j \right)}}(n)}
\end{array}} \right] = g\left( {\left[ {\begin{array}{*{20}{c}}
{{\bf{W}}_{{\rm{in,}}N}^{\left( 1 \right)}}\\
{{\bf{W}}_{{\rm{in,}}N}^{\left( 2 \right)}}\\
 \vdots \\
{{\bf{W}}_{{\rm{in,}}N}^{\left( j \right)}}
\end{array}} \right]{\bf{u}}(n) + } \right.\\
\left. {\left[ {\begin{array}{*{20}{c}}
{{\bf{W}}_{{\rm{r,}}N}^{\left( 1 \right)}}&0&0&0\\
0&{{\bf{W}}_{{\rm{r,}}N}^{\left( 2 \right)}}&0&0\\
 \vdots & \vdots & \ddots & \vdots \\
0&0&0&{{\bf{W}}_{{\rm{r,}}N}^{\left( j \right)}}
\end{array}} \right]\left[ {\begin{array}{*{20}{c}}
{{{\bf{x}}^{\left( 1 \right)}}(n - 1)}\\
{{{\bf{x}}^{\left( 2 \right)}}(n - 1)}\\
 \vdots \\
{{{\bf{x}}^{\left( j \right)}}(n - 1)}
\end{array}} \right] + \left[ {\begin{array}{*{20}{c}}
{{\bf{b}}_N^{\left( 1 \right)}}\\
{{\bf{b}}_N^{\left( 2 \right)}}\\
 \vdots \\
{{\bf{b}}_N^{\left( j \right)}}
\end{array}} \right]} \right)
\end{array},\\
{\bf{Y}}{\rm{ = }}\sum\limits_{k = 1}^j {{\bf{W}}_{{\mathop{\rm out}\nolimits} }^{\left( k \right)}{{\bf{X}}^{\left( k \right)}}},
\end{array}
\end{equation}
\end{scriptsize}where $\mathbf{W}_{\text{in,}N}^{\left( k \right)}$, $\mathbf{W}_{\text{r,}N}^{\left( k \right)}$, $\mathbf{b}_{N}^{\left( k \right)}$, ${{\mathbf{x}}^{\left( k \right)}}(n)$ and ${\bf{W}}_{{\mathop{\rm out}\nolimits} }^{\left( k \right)}$ are the input weight, reservoir weight, bias, reservoir state and output weight of the $k$-th subreservoir, and ${{\mathbf{X}}^{\left( k \right)}}=\left[ {{\mathbf{x}}^{\left( k \right)}}\left( 1 \right),{{\mathbf{x}}^{\left( k \right)}}\left( 2 \right),\ldots ,{{\mathbf{x}}^{\left( k \right)}}\left( {{n}_{\max }} \right) \right]$.

Calculate the current error ${{e}_{0}}:={{e}_{j}}=\mathbf{Y}-\mathbf{T}$. If ${{\left\| e_{j}^{{}} \right\|}_{F}}>\varepsilon $, we need to configure subreservoirs under the supervisory mechanism. Specifically, to prevent over-fitting, an additional condition is introduced for stopping adding subreservoirs and a step size ${{j}_{\text{step}}}$ (${{j}_{\text{step}}}<j$) is used in the early stopping criterion, that is,
\begin{equation}
\label{eq111}
{\left\| {{e_{{\rm{val}},j - {j_{{\rm{step}}}}}}} \right\|_F} \le {\left\| {{e_{{\rm{val}},j - {j_{{\rm{step}}}} + 1}}} \right\|_F} \le  \ldots  \le {\left\| {{e_{{\rm{val}},j}}} \right\|_F},
\end{equation}
where ${{e}_{\text{val},j}}$ represents the validation residual error with $j$ subreservoirs. If (\ref{eq111}) is satisfied, the number of subreservoirs will be set to $j-{{j}_{\text{step}}}$. 

Take $\mathbf{W}_{\text{in,}N}^{\left( j+1 \right)}$, $\mathbf{W}_{\text{r,}N}^{\left( j+1 \right)}$ and $\mathbf{b}_{N}^{\left( j+1 \right)}$ from an adjustable uniform distribution $\left[ -\lambda ,\lambda  \right]$ for ${{G}_{\max }}$ times, and construct the candidates of the subreservoir state $\mathbf{X}_{{}}^{\left( j+1 \right)1},\ldots ,\mathbf{X}_{{}}^{\left( j+1 \right),{{G}_{\max }}}$. 
\begin{remark}
To ensure the echo state property of the model, the feedback matrix $\mathbf{W}_{\text{r,}N}^{\left( j+1 \right)}$ needs to be scaled by
\begin{equation} \label{eq112}
{\bf{W}}_{{\rm{r}},N + 1}^{\left( {j + 1} \right)} \leftarrow \frac{\theta }{{\rho \left( {{\bf{W}}_{{\rm{r}},N + 1}^{\left( {j + 1} \right)}} \right)}}{\bf{W}}_{{\rm{r}},N + 1}^{\left( {j + 1} \right)},
\end{equation}
where ${\rho \left( {{\bf{W}}_{{\rm{r}},N + 1}^{\left( {j + 1} \right)}} \right)}$ is the maximal eigenvalue of $\mathbf{W}_{\text{r,}N}^{\left( j+1 \right)}$ and $\theta $ represents the scaling factor. 
\end{remark}

Seek subreservoirs satisfying the following inequality constraint with an increasing factor $r$, that is,
\begin{equation} \label{eq12}
\begin{array}{l}
{\xi _{j{\rm{ + }}1,q}} = (1 - r - {\mu _{j + 1}})\left\| {e_{j,q}^{}} \right\|_{}^2 - \frac{{{{\left\langle {e_{j,q}^{},{{\bf{X}}^{\left( {j + 1} \right),i}}} \right\rangle }^2}}}{{\left\langle {{{\bf{X}}^{\left( {j + 1} \right),i}},{{\bf{X}}^{\left( {j + 1} \right),i}}} \right\rangle }} \le 0,{\kern 1pt} \\
{\kern 1pt} {\kern 1pt} {\kern 1pt} {\kern 1pt} {\kern 1pt} {\kern 1pt} {\kern 1pt} {\kern 1pt} {\kern 1pt} {\kern 1pt} {\kern 1pt} {\kern 1pt} {\kern 1pt} {\kern 1pt} {\kern 1pt} {\kern 1pt} {\kern 1pt} {\kern 1pt} {\kern 1pt} {\kern 1pt} {\kern 1pt} {\kern 1pt} {\kern 1pt} {\kern 1pt} {\kern 1pt} {\kern 1pt} {\kern 1pt} {\kern 1pt} {\kern 1pt} {\kern 1pt} {\kern 1pt} {\kern 1pt} {\kern 1pt} {\kern 1pt} {\kern 1pt} {\kern 1pt} {\kern 1pt} {\kern 1pt} {\kern 1pt} {\kern 1pt} {\kern 1pt} {\kern 1pt} {\kern 1pt} q = 1,2, \ldots ,L,{\kern 1pt} {\kern 1pt} {\kern 1pt} {\kern 1pt} {\kern 1pt} {\kern 1pt} {\kern 1pt} i = 1,2, \ldots ,{G_{\max }},
\end{array}
\end{equation}
where $L$ denotes the the dimension of the output. A larger positive value $\xi _{j+1}^{{}}\text{=}\sum\limits_{q=1}^{L}{{{\xi }_{j+1,q}}}$ implies the adding subreservoir is better configured. 

The output weight is determined by the least square method, that is,
\begin{equation} \label{eq13}
\begin{array}{*{20}{c}}
{\left[ {{\bf{W}}_{{\mathop{\rm out}\nolimits} }^{\left( 1 \right)}, \ldots ,{\bf{W}}_{{\mathop{\rm out}\nolimits} }^{\left( j+1 \right)}} \right] = \mathop {\arg \min }\limits_{{\bf{W}}_{{\mathop{\rm out}\nolimits} }^{}} \left\| {{\bf{T}} - \sum\limits_{k = 1}^{j+1} {{\bf{W}}_{{\mathop{\rm out}\nolimits} }^{\left( k \right)}{{\bf{X}}^{\left( k \right)}}} } \right\|}
\end{array}.
\end{equation}
Calculate the current residual error $e_{j+1}^{{}}$ and ${{e}_{\text{val},j+1}}$, and update $j=j+1$. Continue adding the subreservoirs until ${\left\| {e_0^{}} \right\|_F} \le \varepsilon $ or $j = J_{\max }^{}$ or (\ref{eq111}) is met, where $J_{\max }^{{}}$ is the maximum number of subreservoirs.

\section{Self-organizing recurrent stochastic configuration networks}
In actual industrial applications, fluctuations in operational variables and system dynamics can lead to nonstationarity. This implies that the statistical characteristics of the data, such as mean, variance, and autocorrelation evolve over time, making it difficult to develop accurate forecasting models. To address this issue, we propose the self-organizing recurrent stochastic configuration networks to learn the meaningful information and patterns from nonstationary data streams. The following section provides detailed insights into the two key phases of the self-organizing learning process: the online update of network parameters and the dynamic adjustment of the network structure.

\subsection{Online update of network parameters}
Online learning enables models to continuously update their parameters as the new data arrives. This section elaborates on the online update of network parameters using the projection algorithm \cite{ref24}.

Consider a BRSCN model expressed by (\ref{eq13}) and let $\mathbf{\hat{g}}\left( n \right)\text{=}{{\left[ {{\mathbf{x}}^{\left( 1 \right)}}\left( n \right),\ldots ,{{\mathbf{x}}^{\left( j \right)}}\left( n \right) \right]}^{\top }}$. Determine ${{\mathbf{W}}_{{\rm out}}}\left( n \right)$ to minimize the following cost function:
\begin{equation} \label{eq15}
\begin{array}{*{20}{c}}
{J = \frac{1}{2}{{\left\| {{{{\mathbf{W}}}_{{\rm out}}}\left( n \right) - {{{\mathbf{W}}}_{{\rm out}}}\left( {n - 1} \right)} \right\|}^2}}\\
{s.t.{\kern 1pt} {\kern 1pt} {\kern 1pt} {\kern 1pt} {\kern 1pt} {\kern 1pt} {\kern 1pt} {\kern 1pt} {\kern 1pt} {\kern 1pt} {\bf{y}}\left( n \right) = {{{\mathbf{W}}}_{{\rm out}}}\left( {n - 1} \right){\bf{\hat g}}(n)}
\end{array}.
\end{equation}
By introducing the Lagrange operator ${{\lambda }_{\text{p}}}$, we have
\begin{equation} \label{eq16}
\begin{array}{l}
{J_e} = \frac{1}{2}{\left\| {{{{\mathbf{W}}}_{\rm out}}\left( n \right) - {{{\mathbf{W}}}_{\rm out}}\left( {n - 1} \right)} \right\|^2}\\
{\kern 1pt} {\kern 1pt} {\kern 1pt} {\kern 1pt} {\kern 1pt} {\kern 1pt} {\kern 1pt} {\kern 1pt} {\kern 1pt} {\kern 1pt} {\kern 1pt} {\kern 1pt} {\kern 1pt} {\kern 1pt} {\kern 1pt} {\kern 1pt}  + {\lambda _{\rm{p}}}\left[ {{\bf{y}}\left( n \right) - {{{\mathbf{W}}}_{\rm out}}\left( {n - 1} \right){\bf{\hat g}}(n)} \right].
\end{array}
\end{equation}
Taking the derivative of (\ref{eq16}) with respect to ${{\mathbf{W}}_{{\rm out}}}\left( n \right)$ and ${{\lambda }_{\text{p}}}$, we have
\begin{equation} \label{eq18}
\left\{ {\begin{array}{*{20}{c}}
{{{{\mathbf{W}}}_{{\rm out}}}\left( n \right) - {{{\mathbf{W}}}_{{\rm out}}}\left( {n - 1} \right) - {\lambda _{\rm{p}}}{{{\bf{\hat g}}}^ \top }(n) = 0}\\
{{\bf{y}}\left( n \right) - {{{\mathbf{W}}}_{{\rm out}}}\left( {n - 1} \right){\bf{\hat g}}(n) = 0}
\end{array}} \right.,
\end{equation}
\begin{equation} \label{eq19}
{\lambda _{\rm{p}}} = \frac{{{\bf{y}}\left( n \right) - {{{\mathbf{W}}}_{{\rm out}}}\left( {n - 1} \right){\bf{\hat g}}(n)}}{{{\bf{\hat g}}{{(n)}^ \top }{\bf{\hat g}}(n)}}.
\end{equation}
Substituting (\ref{eq19}) into (\ref{eq18}), we can obtain
\begin{equation} \label{eq20}
\begin{array}{l}
{{{\mathbf{W}}}_{\rm out}}(n) = {{{\mathbf{W}}}_{\rm out}}(n - 1)  \\
{\kern 1pt} {\kern 1pt} {\kern 1pt} {\kern 1pt} {\kern 1pt} {\kern 1pt} {\kern 1pt} {\kern 1pt} {\kern 1pt} {\kern 1pt} {\kern 1pt} {\kern 1pt} {\kern 1pt} {\kern 1pt} {\kern 1pt} {\kern 1pt} {\kern 1pt} {\kern 1pt} {\kern 1pt} {\kern 1pt} +\frac{{{\bf{\hat g}}{{(n)}^ \top }}}{{{\bf{\hat g}}{{(n)}^ \top }{\bf{\hat g}}(n)}}\left( {{\bf{y}}\left( n \right) - {{{\mathbf{W}}}_{\rm out}}\left( {n - 1} \right){\bf{\hat g}}(n)} \right).
\end{array}
\end{equation}\vspace{-0.5cm}
\begin{figure*}[htbp]
	\begin{center}
		\includegraphics[width=13cm,height=7.5cm]{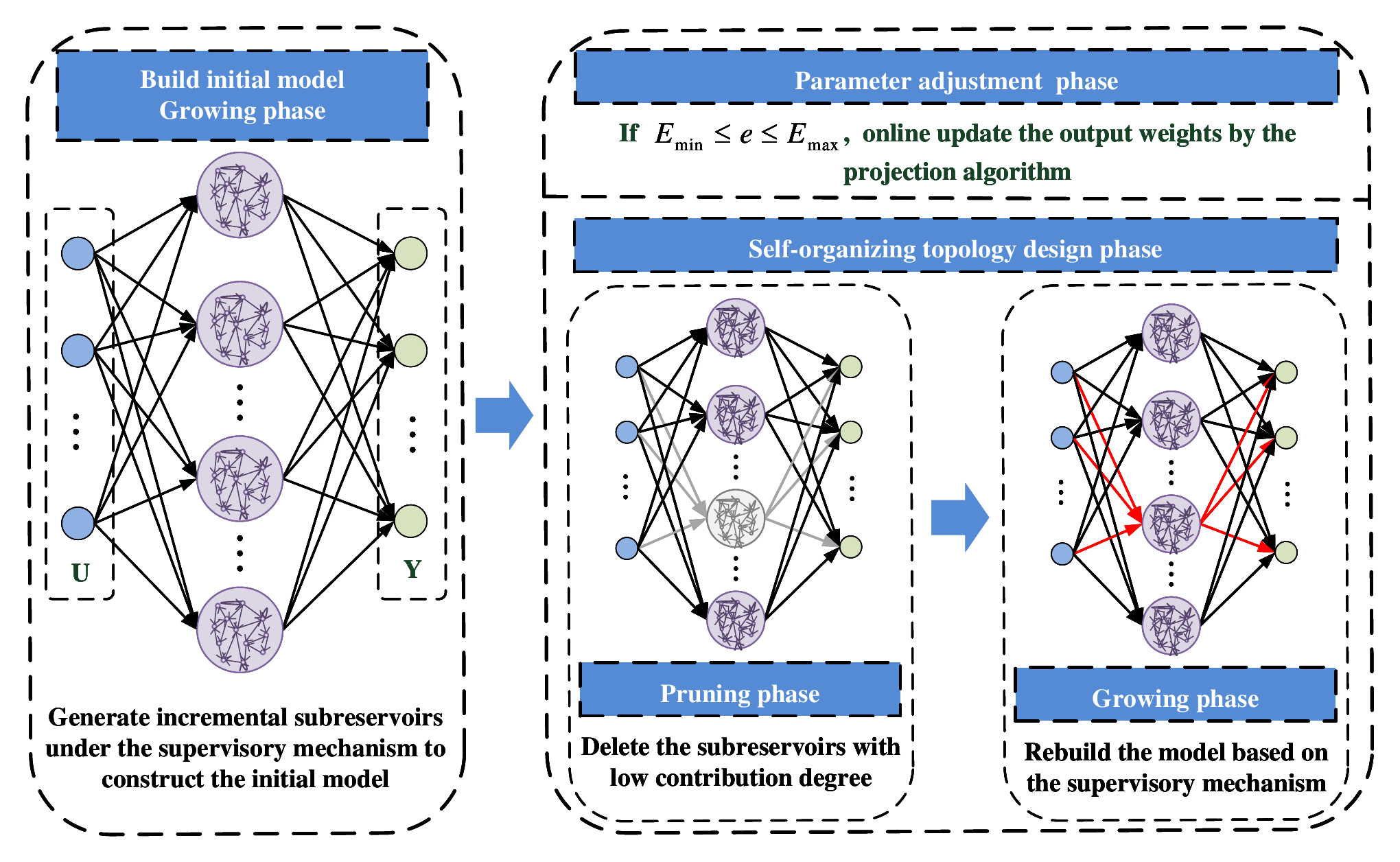}
		\caption{Architecture of the basic SORSCN.}
		\label{fig1}
	\end{center}
 \vspace{-0.5cm}
\end{figure*}

\subsection{Dynamic adjustment of the network structure}
Industrial process data typically exhibit a highly variable distribution. Simply adjusting parameters online is not sufficient to improve the learning capability of the network to accommodate changes in the real-time data streams. To enhance the online self-learning ability of the model in analyzing nonstationary data, we propose a dynamic structure adjustment strategy based on the improved sensitivity analysis and RSC algorithm.

The model output of the BRSCN can be regarded as a linear combination of the states of each subreservoir, that is,
\begin{equation} \label{eq22}
{\bf{Y}}{\rm{ = }}\sum\limits_{k = 1}^J {{\bf{W}}_{{\mathop{\rm out}\nolimits} }^{\left( k \right)}{{\bf{X}}^{\left( k \right)}}},
\end{equation}
where ${{\mathbf{X}}^{\left( k \right)}}\text{=}\left[ {{\mathbf{x}}^{\left( k \right)}}\left( 1 \right),{{\mathbf{x}}^{\left( k \right)}}\left( 2 \right),\ldots ,{{\mathbf{x}}^{\left( k \right)}}\left( {{n}_{\max }} \right) \right]$, $J$ is the total number of subreservoirs. If we prune the $k$-th subreservoir, the output is
\begin{equation} \label{eq23}
\begin{array}{l}
{\bf{Y'}}{\rm{ = }}{\bf{W}}_{{\mathop{\rm out}\nolimits} }^{\left( 1 \right)}{{\bf{X}}^{\left( 1 \right)}}{\rm{ + }} \cdots {\rm{ + }}{\bf{W}}_{{\mathop{\rm out}\nolimits} }^{\left( {k - 1} \right)}{{\bf{X}}^{\left( {k - 1} \right)}}\\
{\kern 1pt} {\kern 1pt} {\kern 1pt} {\kern 1pt} {\kern 1pt} {\kern 1pt} {\kern 1pt} {\kern 1pt} {\kern 1pt} {\kern 1pt} {\kern 1pt} {\kern 1pt} {\kern 1pt} {\kern 1pt} {\kern 1pt} {\kern 1pt} {\kern 1pt} {\kern 1pt} {\kern 1pt} {\kern 1pt} {\rm{ + }}{\bf{W}}_{{\mathop{\rm out}\nolimits} }^{\left( {k + 1} \right)}{{\bf{X}}^{\left( {k + 1} \right)}} \cdots {\rm{ + }}{\bf{W}}_{{\mathop{\rm out}\nolimits} }^{\left( J \right)}{{\bf{X}}^{\left( J \right)}}.
\end{array}
\end{equation}
The sensitivity of the model residual error to the $k$-th subreservoir is denoted as
\begin{equation} \label{eq24}
{S_k}{\rm{ = }}\frac{1}{{{n_w}}}\sum\limits_{n = {n_h} + 1}^{{n_h} + {n_w}} {\left\| {{\bf{W}}_{{\mathop{\rm out}\nolimits} }^{\left( k \right)}{{\bf{x}}^{\left( k \right)}}\left( n \right)} \right\|} ,
\end{equation}
where ${{n}_{h}}$ and ${{n}_{w}}$ are the number of historical and arriving samples, respectively. A larger ${{S}_{k}}$ indicates that the $k$-th subreservoir contributes more to the model output. Thus, the contribution of the subreservoir can be ranked as ${{S}_{1}}^{\prime }>{{S}_{2}}^{\prime }>\cdots >{{S}_{J}}^{\prime }$. The model scale adaptability (MSA) can be defined based on the sensitivity, that is,
\begin{equation} \label{eq25}
{M_{{J_K}}}{\rm{ = }}\frac{{\sum\limits_{k = 1}^{{J_K}} {{{S'}_k}} }}{{\sum\limits_{k = 1}^J {{{S'}_k}} }},1 \le {J_K} \le J.
\end{equation}
The optimal number of subreservoirs ${{J}_{M}}$ can be determined by MSA and we have
\begin{equation} \label{eq26}
{J_M}{\rm{ = min}}\left\{ {{J_K}|{M_{{J_K}}} \ge \gamma ,1 \le {J_K} \le J} \right\}{\rm{,}}
\end{equation}
where $\gamma \left( 0\le \gamma \le 1 \right)$ is the threshold of MSA, ${{J}_{M}}$ is the number of reserved subreservoirs, and $J-{{J}_{M}}$ is the number of redundant subreservoirs. 

Specifically, we present an improved MSA, where the sensitivity analysis with correlation metric is used to prune the highly correlated subreservoirs and enhance the learning ability of the model. Then, (\ref{eq25}) can be rewritten as
\begin{equation} \label{eq27}
{M_{{J_K}}}{\rm{ = }}\frac{{\sum\limits_{k = 1}^{{J_K}} {{{S'}_k}} }}{{\sum\limits_{k = 1}^J {{{S'}_k}} }} + \alpha \frac{{\sum\limits_{k = 1}^{{J_K}} {{C_k}} }}{{\sum\limits_{k = 1}^J {{C_k}} }},1 \le {J_K} \le J,
\end{equation}
where $\alpha $ is a weighted coefficient. ${{C}_{k}}$ is given by
\begin{equation} \label{eq28}
\left\{ {\begin{array}{*{20}{c}}
{{C_k} = 1 - \frac{{{c_k}}}{{{c_1} + {c_2} +  \cdots {c_J}}}}\\
{{c_k} = \sum\limits_{k' = 1}^J {\left| {corrcoef\left( {{{{\bf{X'}}}^{\left( k \right)}},{{{\bf{X'}}}^{\left( {k'} \right)}}} \right)} \right|} ,k' \ne k}
\end{array}} \right.,
\end{equation}
where $corrcoef\left( \bullet  \right)$ is the function of correlation metric, ${{\mathbf{{X}'}}^{\left( k \right)}}=\left[ {{\mathbf{x}}^{\left( k \right)}}\left( {{n}_{h}}+1 \right),{{\mathbf{x}}^{\left( k \right)}}\left( {{n}_{h}}+2 \right),\ldots ,{{\mathbf{x}}^{\left( k \right)}}\left( {{n}_{h}}+{{n}_{w}} \right) \right]$. 

Suppose that we have pruned $J-{{J}_{M}}$ subreservoirs, the model output can be calculated by
\begin{equation} \label{eq29}
F_{{J_M}}^{} = \sum\limits_{k = 1}^{{J_M}} {{\bf{W}}_{{\mathop{\rm out}\nolimits} }^{\left( k \right)}{{{\bf{X'}}}^{\left( k \right)}}} .
\end{equation}
The residual error is ${{{e}'}_{{{J}_{M}}}}=\left[ {{{{e}'}}_{{{J}_{M}},1}},{{{{e}'}}_{{{J}_{M}},2}},\ldots ,{{{{e}'}}_{{{J}_{M}},L}} \right]$. If ${{\left\| {{{{e}'}}_{{{J}_{M}}}} \right\|}_{F}}$ is larger than the upper bound of the preset error interval, the network needs to add new subreservoirs which satisfy the following inequality constraint:
\begin{equation} \label{eq30}
(1 - r - {\mu _{{J_M} + 1}})\left\| {{{e'}_{{J_M},q}}} \right\|_{}^2 - \frac{{{{\left\langle {{{e'}_{{J_M},q}},{{{\bf{X'}}}^{\left( {{J_M} + 1} \right)}}} \right\rangle }^2}}}{{\left\langle {{{{\bf{X'}}}^{\left( {{J_M} + 1} \right)}},{{{\bf{X'}}}^{\left( {{J_M} + 1} \right)}}} \right\rangle }} \le 0.
\end{equation}
Finally, the output weight can be calculated by the least square method, that is, 
\begin{small}
 \begin{equation} \label{eq31}
\begin{array}{l}
\left[ {{\bf{W}}_{{\mathop{\rm out}\nolimits} }^{\left( 1 \right)*}, \ldots ,{\bf{W}}_{{\mathop{\rm out}\nolimits} }^{\left( {{J_M} + 1} \right)*}} \right] = \mathop {\arg \min }\limits_{{\bf{W}}_{{\mathop{\rm out}\nolimits} }^{}} \left\| {{\bf{T'}} - \sum\limits_{k = 1}^{{J_M} + 1} {{\bf{W}}_{{\mathop{\rm out}\nolimits} }^{\left( k \right)}{{{\bf{X'}}}^{\left( k \right)}}} } \right\|,
\end{array}
\end{equation}   
\end{small}where $\mathbf{{T}'}$ is the desired output of the newly arriving samples. Calculate the current residual error. If the residual error is smaller than the error threshold, the construction is complete, otherwise, continue configuring subreservoirs until meeting the terminal conditions.
\subsection{Algorithm description}
For actual industrial processes, the evolving data distribution poses significant challenges for the well-trained models. To address this problem, we propose the self-organizing RSCNs for modelling nonlinear systems with nonstationary data streams. From the algorithmic perspective, the implementation of SORSCNs involves three phases. The first phase termed the growing phase, is dedicated to constructing the initial network model and capturing crucial patterns within the historical data. Subsequently, the pruning phase aims to enhance the model's efficiency by eliminating subreservoirs that do not significantly contribute to the learning task. The second growing phase allows the model to adjust the network structure based on the newly arriving data. By incorporating the growing-pruning-growing phases, the model's continuous learning capability for unknown nonstationary dynamics is improved while maintaining a streamlined and optimized network structure. The architecture of the basic SORSCN is shown in Fig. \ref{fig1}, and the construction process of the SORSCN can be summarized as follows.

Step 1: Set an appropriate time window and error interval $\left[ {{E}_{\min }},{{E}_{\max }} \right]$ according to the data generated from the nonlinear system. Establish an initial BRSCN model based on the historical data.

Step 2: Input the data from the new time window into the initial network and calculate the model output and the corresponding residual error $e$.

Step 3: If $e$ is within the error interval $\left[ {{E}_{\min }},{{E}_{\max }} \right]$, the network parameters are updated online by using the projection algorithm.

Step 4: If $e$ is larger than the upper bound ${{E}_{\max }}$ of the error interval, the network structure is dynamically adjusted based on the improved sensitivity analysis and the RSC algorithm until satisfying the terminal conditions.

Step 5: Return to Step 2 to perform the self-organizing learning for the newly arriving data streams.

\begin{algorithm}[t]\footnotesize
	\caption{Self-organizing RSC (SORSC)}\label{algo1}	
	\KwIn{Given the datasets $\left\{ \left( \mathbf{u}(n),\mathbf{y}\left( n \right) \right)|\mathbf{u}(n)\in {{\mathbb{R}}^{K}},\mathbf{y}\left( n \right)\in {{\mathbb{R}}^{L}} \right\}$, $n=1,2,\ldots ,{{n}_{h}}$, the size of the subreservoir $N$, the maximum number of subreservoirs $J_{\max }^{{}}$, training error threshold $\varepsilon $, random parameter scalars $\mathbf{\gamma }\text{=}\left\{ {{\lambda }_{1}},...,{{\lambda }_{\max }} \right\}$, contractive factors $\left\{ {{r_1}, \ldots ,{r_t}} \right\},(0 < {r_i} < 1)$, and the maximum number of stochastic configurations ${{G}_{\max }}$.}
    \KwOut{SORSCN}
    \tcp{\textbf{Phase 1: Construct the initial BRSCN model}}
     Initialize network parameters;\\
    Build a block incremental recurrent stochastic configuration network based on (\ref{eq13});\\ 
    Return the trained input weights $\mathbf{\mathbf{W}}_{\text{in}}^{{}}$, biases $\mathbf{\mathbf{b}}$, reservoir weights $\mathbf{\mathbf{W}}_{\text{r}}^{{}}$, and output weights $\mathbf{\mathbf{W}}_{\operatorname{out}}^{{}}$ of the $j$ subreservoirs.\\
    \tcp{\textbf{Phase 2: Online self-organizing}}
    Input the real-time data $\left\{ \mathbf{u}({{n}_{h}}+1),\ldots ,\mathbf{u}({{n}_{h}}+{{n}_{w}}) \right\}$, and corresponding output $\left\{ \mathbf{y}({{n}_{h}}+1),\ldots ,\mathbf{y}({{n}_{h}}+{{n}_{w}}) \right\}$, the trained input weights $\mathbf{\mathbf{W}}_{\text{in}}^{{}}$, biases $\mathbf{\mathbf{b}}$, reservoir weights $\mathbf{\mathbf{W}}_{\text{r}}^{{}}$, output weights $\mathbf{\mathbf{W}}_{\operatorname{out}}^{{}}$, error interval $\left[ {{E}_{\min }},{{E}_{\max }} \right]$, the threshold of model scale adaptability $\gamma $, and the weighted coefficient $\alpha $.\\
    Calculate the model output and current residual error $e$;\\
            \If {${{E}_{\min }}\le e\le {{E}_{\max }}$}
            {Calculate the reservoir state matrix $\mathbf{\hat{g}}\left( n \right)\text{=}{{\left[ {{\mathbf{x}}^{\left( 1 \right)}}\left( n \right),{{\mathbf{x}}^{\left( 2 \right)}}\left( n \right),\ldots ,{{\mathbf{x}}^{\left( j \right)}}\left( n \right) \right]}^{\top }}$ based on (\ref{eq13});\\
            Update $\mathbf{\mathbf{W}}_{\operatorname{out}}^{{}}\left( n \right)$ based on (\ref{eq20});\\
             Return $\mathbf{\mathbf{W}}_{\operatorname{out}}^{{}}\left( n \right)$;}
            \If{$e>{{E}_{\max }}$}
                {\For {$k=1,2,\ldots ,J$,}{
                     Calculate the sensitivity of each subreservoir ${{S}_{k}}$ based on (\ref{eq24}) and rank them as ${{S}_{1}}^{\prime }>{{S}_{2}}^{\prime }>\cdots >{{S}_{J}}^{\prime }$;\\
                    Calculate the improved model scale adaptability based on (\ref{eq27}) and (\ref{eq28});\\
                }
                Determine the optimal number of subreservoirs ${{J}_{M}}$;\\
                Prune the network structure and retain the first ${{J}_{M}}$ subreservoirs with the high contribution degree;\\
                \tcp{\textbf{(Rebuild the pruned network based on the supervisory mechanism)}}
                \While {${{J}_{M}}\le {{J}_{\max }}$ and ${{\left\| e \right\|}_{F}}>\varepsilon $}{
                        Randomly assign the input weights, biases, and reservoir weights from the adjustable uniform distribution $\left[ -\lambda ,\lambda  \right]$, and construct candidate subreservoirs;\\
                        Find the optimal  $\mathbf{W}_{\text{in,}N}^{\left( {{J}_{M}}+1 \right)*}$, $\mathbf{W}_{\text{r,}N}^{\left( {{J}_{M}}+1 \right)*}$, and $\mathbf{b}_{N}^{\left( {{J}_{M}}+1 \right)*}$ that satisfy the inequality constraint in (\ref{eq30});\\
                        Calculate the output weights $\mathbf{W}_{\operatorname{out}}^{*}$ based on (\ref{eq31});\\
                        Calculate ${{{e}'}_{{{J}_{M}}+1}}=e-\mathbf{W}_{\operatorname{out}}^{\left( {{J}_{M}}+1 \right)*}{{\mathbf{{X}'}}^{\left( {{J}_{M}}+1 \right)*}}$;\\
                        Update $e:={{{e}'}_{{{J}_{M}}+1}}$, and ${{J}_{M}}={{J}_{M}}+1$;\\
                    }
                }
            Return ${{\bf{W}}_{{\mathop{\rm out}\nolimits} }^{\left( 1 \right)*}, \ldots ,{\bf{W}}_{{\mathop{\rm out}\nolimits} }^{\left( {{J_M} + 1} \right)*}}$, $\mathbf{W}_{\text{in,}N}^{\left( 1 \right)*},\ldots ,\mathbf{W}_{\text{in,}N}^{\left( {{J}_{M}} \right)*}$, $\mathbf{W}_{\text{r,}N}^{\left( 1 \right)*},\ldots ,\mathbf{W}_{\text{r,}N}^{\left( {{J}_{M}} \right)*}$, and $\mathbf{b}_{N}^{\left( 1 \right)*},\ldots ,\mathbf{b}_{N}^{\left( {{J}_{M}} \right)*}$.
\end{algorithm}

\section{Experiment results}
This section reports the experimental results on the two industrial data predictive analyses. Comparisons among the original ESNs, RSCNs, online self-learning SCNs (OSL-SCNs) \cite{ref17}, self-organizing modular ESNs (SOMESNs) \cite{ref20}, and our proposed SORSCNs are carried out. Specifically, models using the MSA defined by (\ref{eq25}) and (\ref{eq27}) are named SORSCN1 and SORSCN2, respectively. The normalized root means square error (NRMSE) is used to evaluate the model performance, that is,  
\begin{equation} \label{eq33}
NRMSE=\sqrt{\frac{\sum\limits_{n=1}^{{{n}_{max}}}{{{\left( \mathbf{y}\left( n \right)-\mathbf{t}\left( n \right) \right)}^{2}}}}{{{n}_{max}}\operatorname{var}\left( \mathbf{t} \right)}},
\end{equation}
where $\operatorname{var}\left( \mathbf{t} \right)$ is the variance of the desired output.

The parameter settings are as follows. The reservoir sparsity and scaling factor of spectral radius are set to $\left[ 0.01,0.03 \right]$ and $\left[ 0.5,1 \right]$. The scope setting of input and reservoir weights for the ESN and SOMESN is set as $\lambda \in \left[ 0.1,1 \right]$. Specifically, for networks with modular adjustment structure, the subreservoir size is set to 10. The RSC frameworks are built with weight scale sequence $\left\{ 0.5,1,5,10,30,50,100 \right\}$, contractive sequence $r=\left[ 0.9,0.99,0.999,0.9999,0.99999 \right]$, the maximum number of stochastic configurations ${{G}_{\max }}=100$, and training tolerance ${{\varepsilon }}={{10}^{-5}}$. The initial reservoir size in RSCN is set to 5. The grid search method is used to determine the hyperparameters, including the scaling factor $\theta $ in (\ref{eq112}) and the reservoir size $N$. The model performance is evaluated through the mean and standard deviation of the NRMSE, derived from 50 independent trials conducted under the same conditions for each experiment.

\subsection{Soft sensing of the butane concentration in the dehumanizer column process}
\begin{figure}[htbp]
	\begin{center}
		\includegraphics[width=6cm]{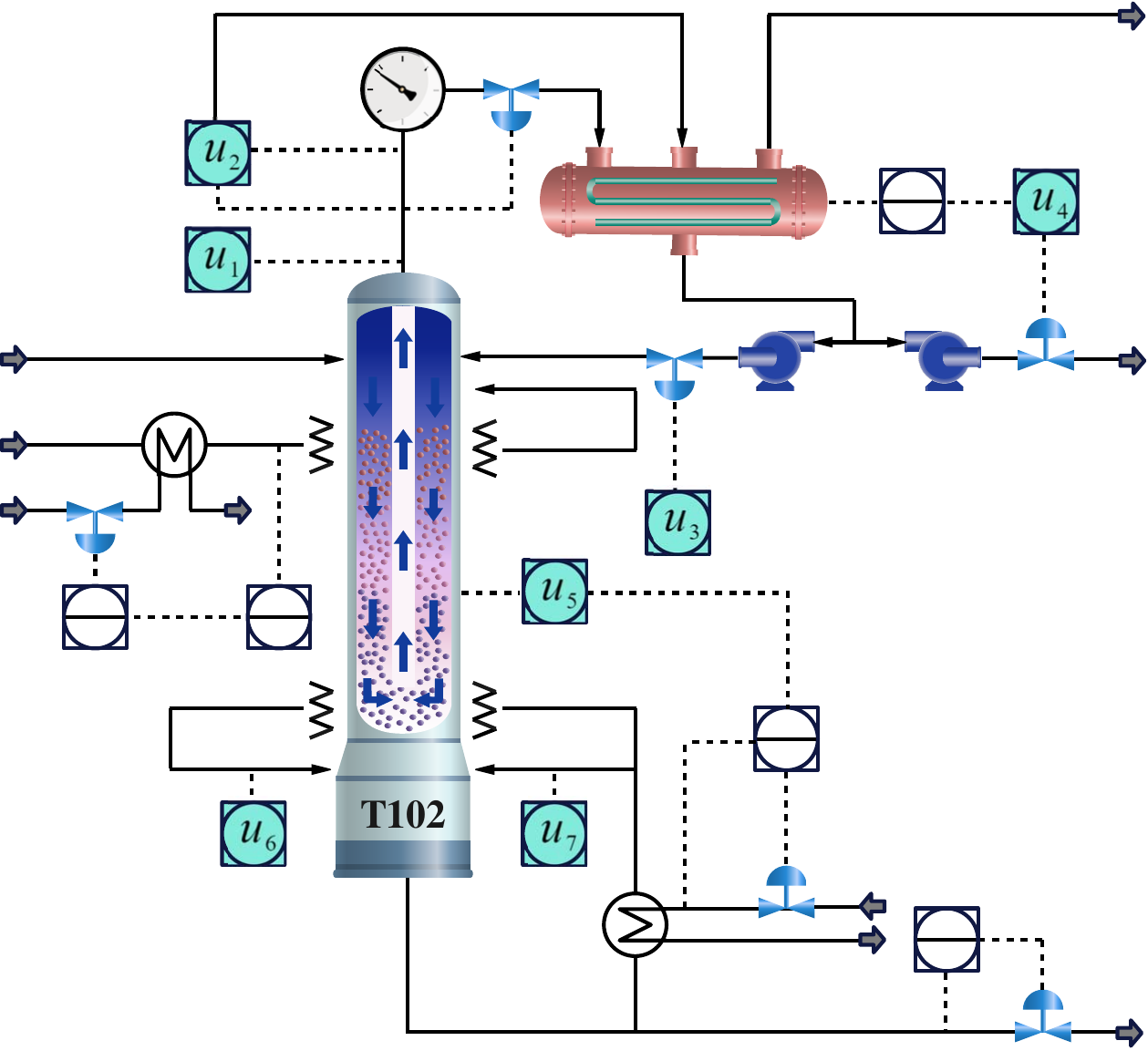}
		\caption{Flowchart of the debutanizer column process.}
		\label{fig3}
	\end{center}
 \vspace{-0.3cm}
\end{figure}

The dehumanizer column plays a crucial role in petroleum refining, primarily used for desulfurization and naphtha cracking. The specific operational process is illustrated in Fig.~\ref{fig3}. This refining process involves seven key variables, including tower top temperature ${{u}_{1}}$, tower top pressure ${{u}_{2}}$, tower top reflux flow ${{u}_{3}}$, tower top product outflow ${{u}_{4}}$, 6-th tray temperature ${{u}_{5}}$, tower bottom temperature ${{u}_{6}}$, and tower bottom pressure ${{u}_{7}}$. One dominant variable $y$ is the concentration of butane at the bottom of the tower, which must adhere to quality control standards by minimizing its levels. In \cite{ref26}, a well-designed combination of variables was presented to obtain the butane concentration $y$, that is, \vspace{-0.3cm}

\begin{small}
    \begin{equation} \label{eq39}
\begin{array}{l}
y\left( n \right) = f\left( {{u_1}\left( n \right),} \right.{u_2}\left( n \right),{u_3}\left( n \right),{u_4}\left( n \right),{u_5}\left( n \right),{u_5}\left( {n - 1} \right),\\
{\kern 1pt} {\kern 1pt} {\kern 1pt} {\kern 1pt} {\kern 1pt} {\kern 1pt} {\kern 1pt} {\kern 1pt} {\kern 1pt} {\kern 1pt} {\kern 1pt} {\kern 1pt} {\kern 1pt} {\kern 1pt} {\kern 1pt} {\kern 1pt} {\kern 1pt} {\kern 1pt} {\kern 1pt} {\kern 1pt} {\kern 1pt} {\kern 1pt} {\kern 1pt} {\kern 1pt} {\kern 1pt} {\kern 1pt} {\kern 1pt} {\kern 1pt} {\kern 1pt} {\kern 1pt} {\kern 1pt} {\kern 1pt} {\kern 1pt} {\kern 1pt} {\kern 1pt} {\kern 1pt} {\kern 1pt} {\kern 1pt} {\kern 1pt} {\kern 1pt} {\kern 1pt} {\kern 1pt} {\kern 1pt} {\kern 1pt} {\kern 1pt} {\kern 1pt} {\kern 1pt} {\kern 1pt} {\kern 1pt} {\kern 1pt} {u_5}\left( {n - 2} \right),{u_5}\left( {n - 3} \right),\left( {{u_1}\left( n \right) + {u_2}\left( n \right)} \right)/2,\\
{\kern 1pt} {\kern 1pt} {\kern 1pt} {\kern 1pt} {\kern 1pt} {\kern 1pt} {\kern 1pt} {\kern 1pt} {\kern 1pt} {\kern 1pt} {\kern 1pt} {\kern 1pt} {\kern 1pt} {\kern 1pt} {\kern 1pt} {\kern 1pt} {\kern 1pt} {\kern 1pt} {\kern 1pt} {\kern 1pt} {\kern 1pt} {\kern 1pt} {\kern 1pt} {\kern 1pt} {\kern 1pt} {\kern 1pt} {\kern 1pt} {\kern 1pt} {\kern 1pt} {\kern 1pt} {\kern 1pt} {\kern 1pt} {\kern 1pt} {\kern 1pt} {\kern 1pt} {\kern 1pt} {\kern 1pt} {\kern 1pt} {\kern 1pt} {\kern 1pt} {\kern 1pt} {\kern 1pt} {\kern 1pt} {\kern 1pt} {\kern 1pt} {\kern 1pt} {\kern 1pt} \left. {{\kern 1pt} {\kern 1pt} {\kern 1pt} y\left( {n - 1} \right),y\left( {n - 2} \right),y\left( {n - 3} \right),y\left( {n - 4} \right)} \right).
\end{array}
\end{equation}
\end{small}Considering the order uncertainty, the input is set as $\left[ {{u}_{1}}\left( n \right), \right.{{u}_{2}}\left( n \right),{{u}_{3}}\left( n \right),{{u}_{4}}\left( n \right),$$\left. {{u}_{5}}\left( n \right),y\left( n-1 \right) \right]$ in our experiments. The data for the naphtha cracking process is collected in real-time, comprising a total of 2394 samples. Samples of 1-1500 steps are used to train the network, the following 894 samples are used for testing. The Gaussian white noise is added to the testing set to generate the validation set. The initial 100 samples of each set are washed out. 

\begin{figure}[htbp]
	\begin{center}
		\includegraphics[width=9cm]{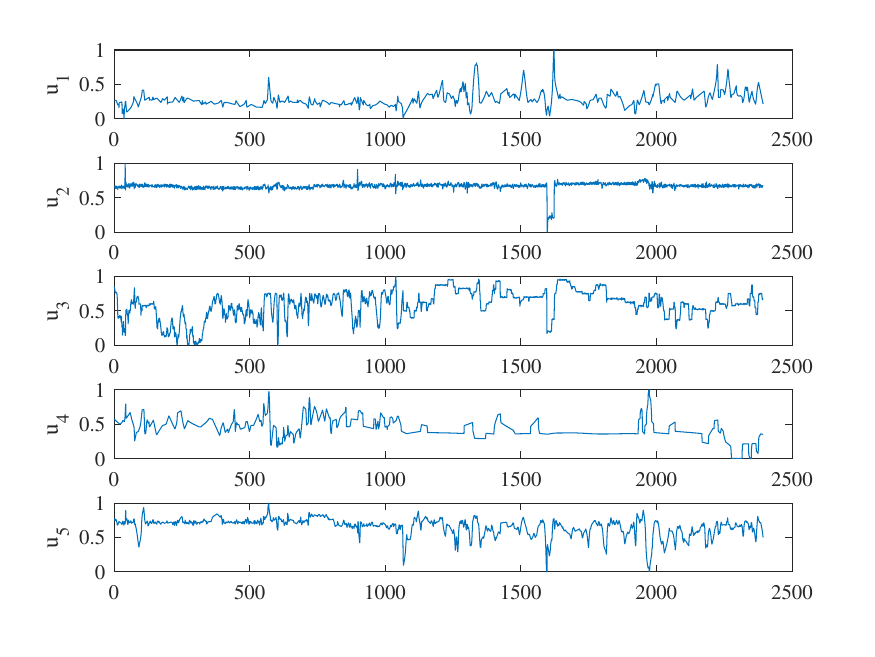}
		\caption{Distribution of the five input variables (${u}_{1}$-${u}_{5}$) for the debutanizer column process.}
		\label{fig311}
	\end{center}
\end{figure}

Fig.~\ref{fig311} shows the the distribution changes of various input variables for the debutanizer column process. These variables exhibit significant nonstationary characteristics, with no consistent cyclical patterns in their fluctuations. Such complex dynamics are typical in real-world industrial processes, making them difficult to model or predict using traditional periodic analysis methods.\vspace{-0.5cm}

\begin{figure}[htpb]
	\centering
	\subfloat[Prediction results]{\includegraphics[width=9cm]{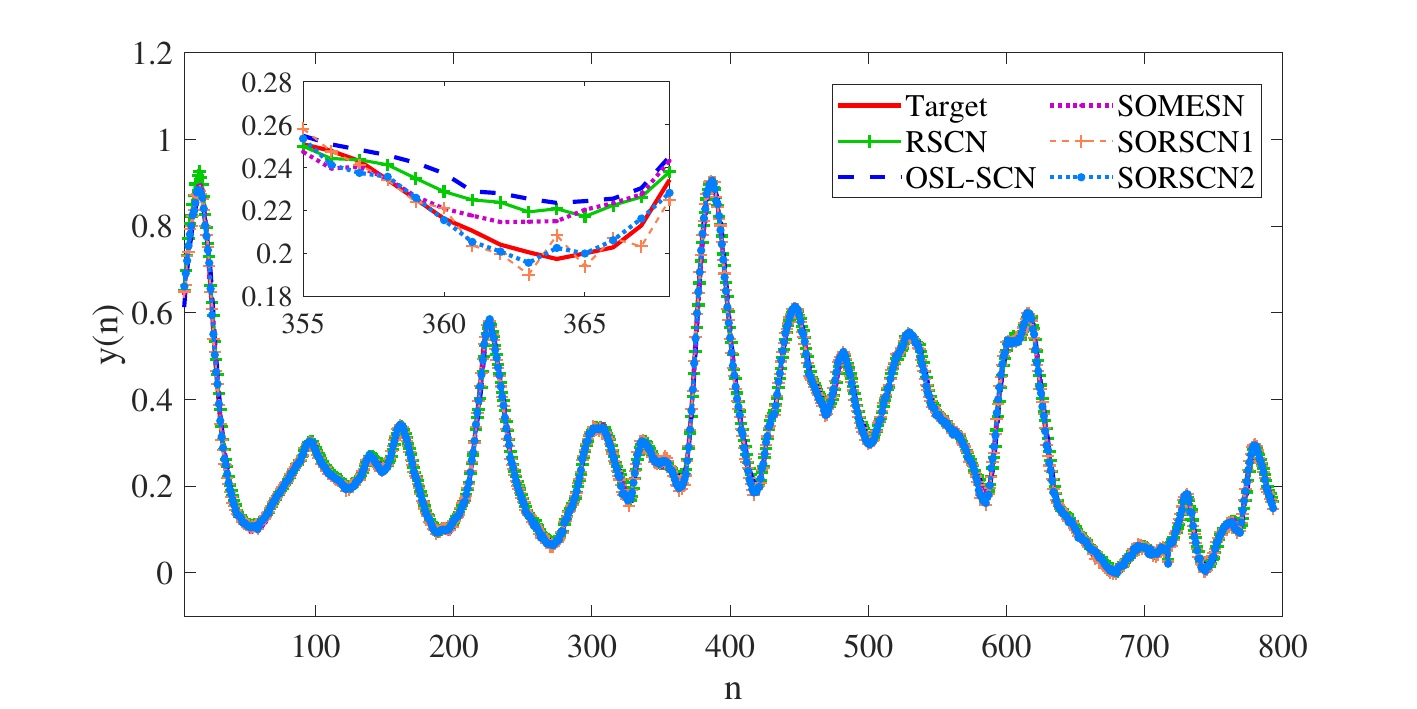}}\\
    \vspace{-0.3cm}
	\subfloat[Prediction errors]{\includegraphics[width=9cm]{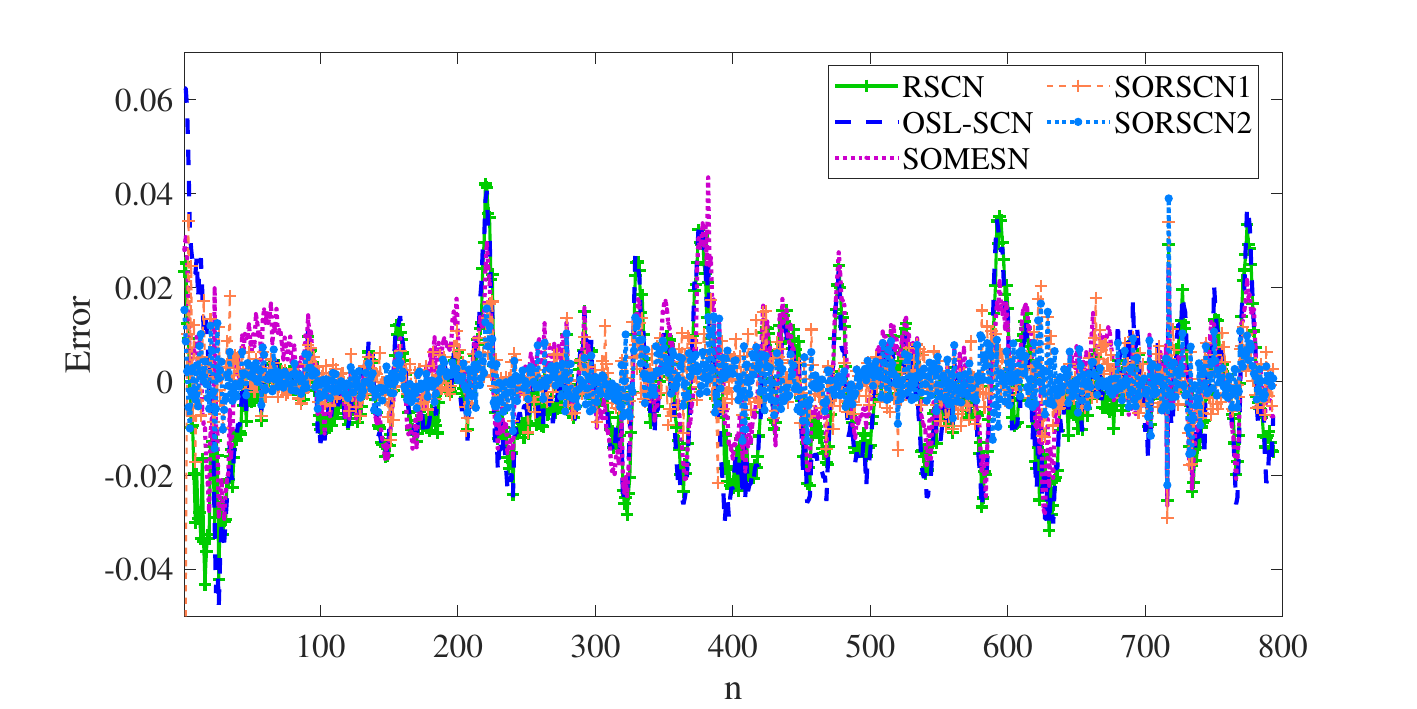}}
	\caption{Prediction fitting curves and error values of different models for the debutanizer column.}
	\label{fig4}
\end{figure}

\setlength{\abovecaptionskip}{-0.5cm}
\vspace{-0.5cm}
Fig.~\ref{fig4} illustrates the prediction fitting curves and error values of ESN, RSCN, and the proposed SORSCNs on the debutanizer column process. The output of SORSCN2 closely aligns with the target output, resulting in smaller prediction errors at each step compared to the other models. This outcome underscores the superior self-organizing capability of the proposed method, which allows it to dynamically adapt to evolving data, thereby addressing the challenges of nonstationary systems more effectively than the original ESN and RSCN.
\setlength{\abovecaptionskip}{5pt}

\begin{figure}[htbp]
\vspace{-0.5cm}
	\begin{center}
		\includegraphics[width=8cm]{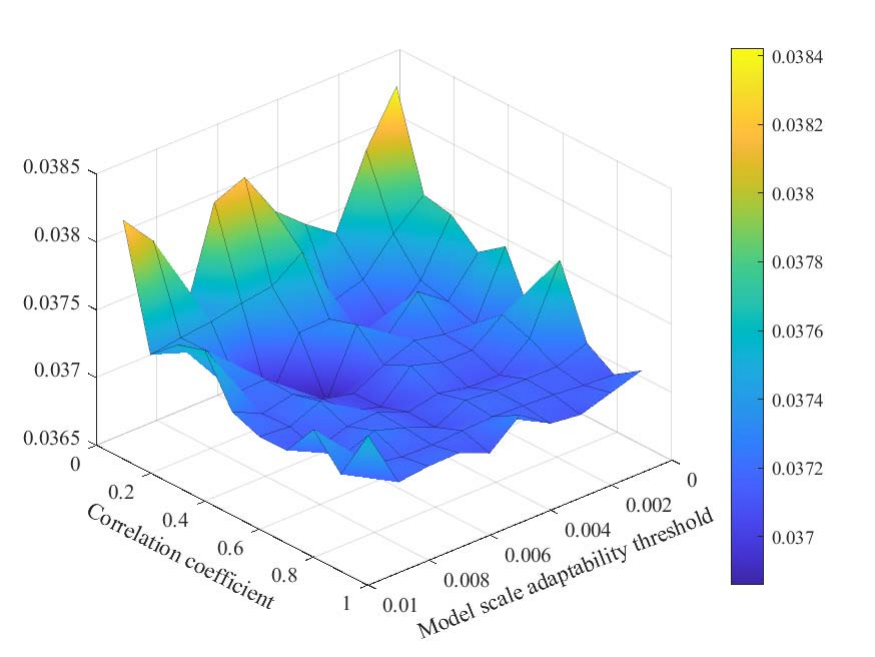}
		\caption{Testing NRMSE of SORSCN2 with different parameters (correlation coefficient $\alpha $  and the threshold of model scale adaptability $\gamma $ ) on the debutanizer column process.}
		\label{fig411}
	\end{center}
 \vspace{-0.7cm}
\end{figure}

Fig.~\ref{fig411} shows the impact of correlation coefficient $\alpha $ and the threshold of model scale adaptability $\gamma $ on the testing performance. It can be seen that a small $\alpha$ may lead to poor prediction accuracy, and the best result can be obtained when $\alpha=0.5$ and $\gamma=0.006$. These findings suggest that incorporating the correlation metric into sensitivity analysis can enhance the model's learning capability and achieve sound generalization performance.\vspace{-0.2cm}

\subsection{Short-term power load forecasting}
\begin{figure}[htbp]
	\begin{center}
		\includegraphics[width=8cm]{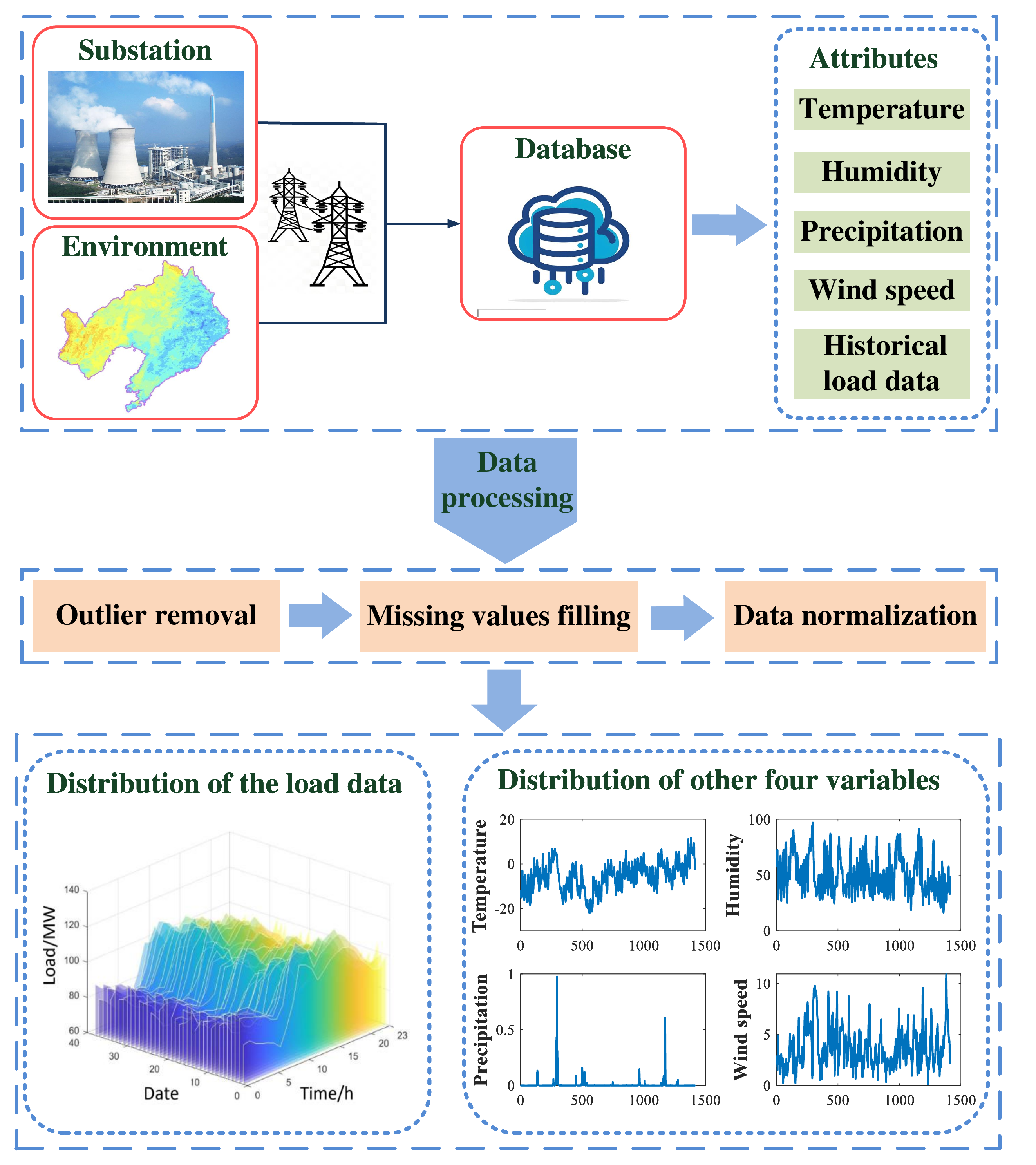}
		\caption{Flowchart of the data collection and processing for the short-term power load forecasting.}
		\label{fig5}
	\end{center}
 \vspace{-0.6cm}
\end{figure}
\begin{figure}[htbp]
 \vspace{-0.3cm}
	\begin{center}
		\includegraphics[width=9cm]{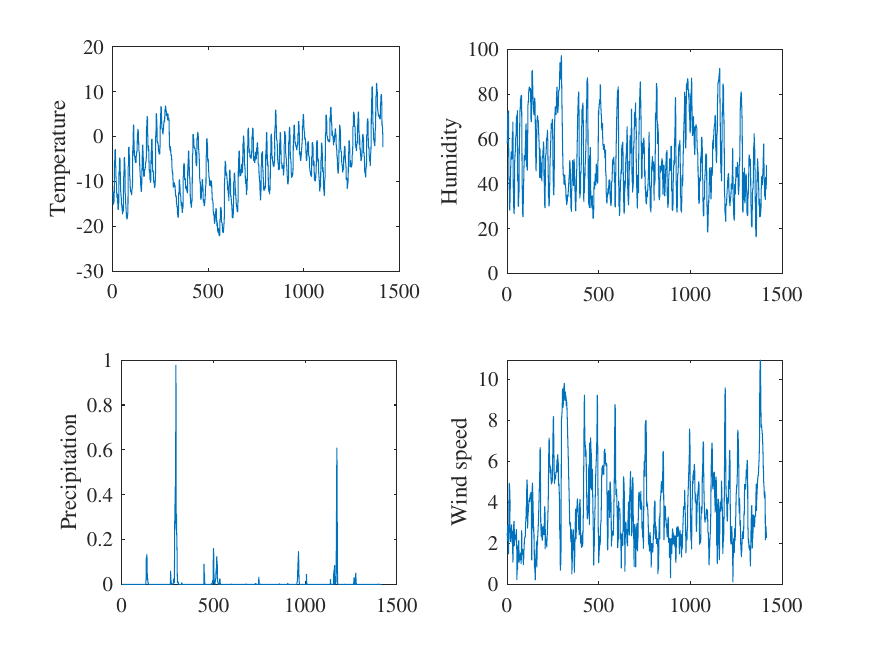}
		\caption{Distribution of the four input variables for the short-term power load forecasting.}
		\label{fig511}
	\end{center}
 \vspace{-0.3cm}
\end{figure}

Short-term power load forecasting is a critical aspect of ensuring the reliability and efficiency of power systems while also reducing costs. This research delves into an analysis of load data obtained from a 500kV substation located in Liaoning Province, China. The dataset is collected hourly over a period spanning January to February 2023, totalling 59 days. Various environmental factors such as temperature ${u_1}$, humidity ${u_2}$, precipitation ${u_3}$, and wind speed ${u_4}$ are taken into account to forecast power load $y$. The process of data collection and analysis for short-term electricity load forecasting is illustrated in Fig.~\ref{fig5}. The distribution changes of the variables are shown in Fig.~\ref{fig511}, where each variable exhibits irregular fluctuations over time, demonstrating clear nonstationary characteristics. The dataset comprises of 1415 samples, with 1000 allocated for training and 415 for testing. Gaussian noise is introduced to the testing set to establish the validation set. Considering order uncertainty, $\left[ {{u}_{1}}\left( n \right), \right.$$\left. {{u}_{2}}\left( n \right),{{u}_{3}}\left( n \right),{{u}_{4}}\left( n \right),y\left( n-1 \right) \right]$ is utilized to predict $y\left( n \right)$ in our experiments. The first 30 samples from each set are washed out.
\setlength{\abovecaptionskip}{-0.3cm}

\begin{figure}[htpb]
\vspace{-0.3cm}
	\centering
	\subfloat[Prediction results]{\includegraphics[width=9cm]{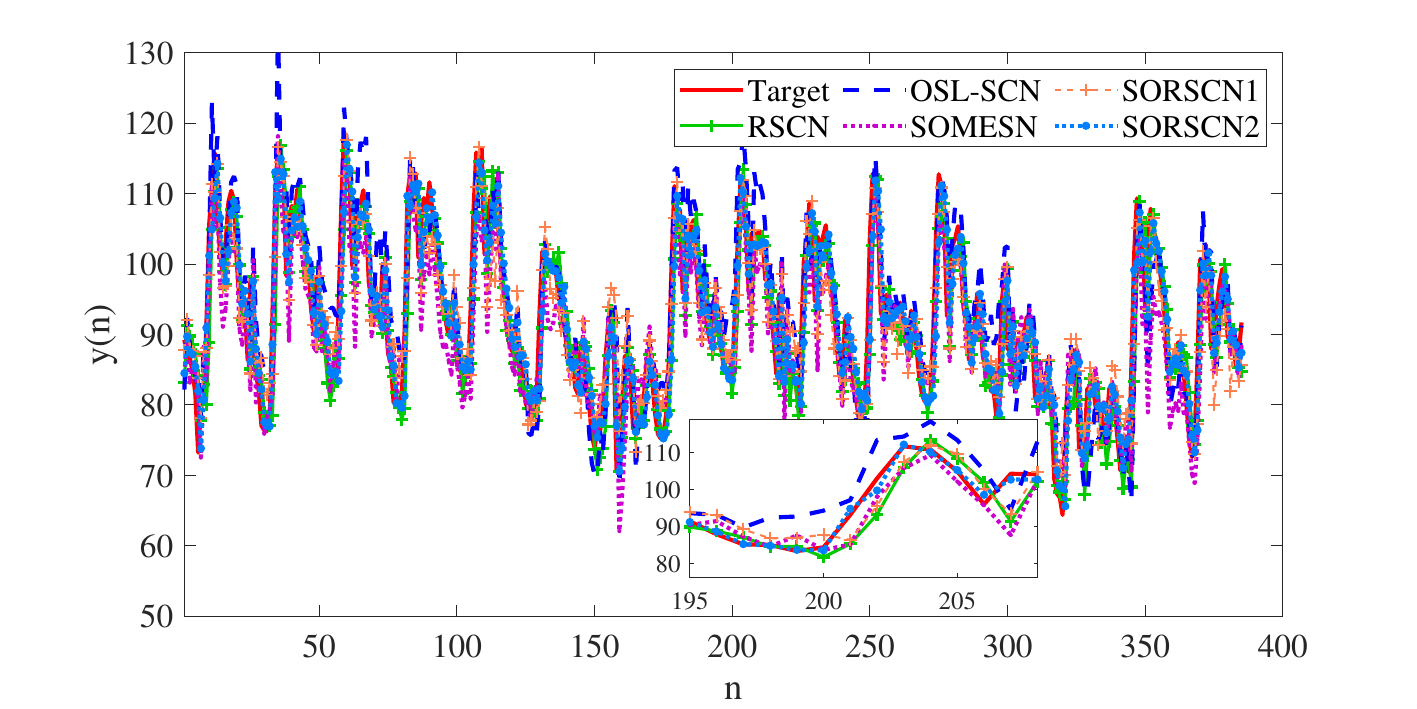}}\\
    \vspace{-0.3cm}
	\subfloat[Prediction errors]{\includegraphics[width=9cm]{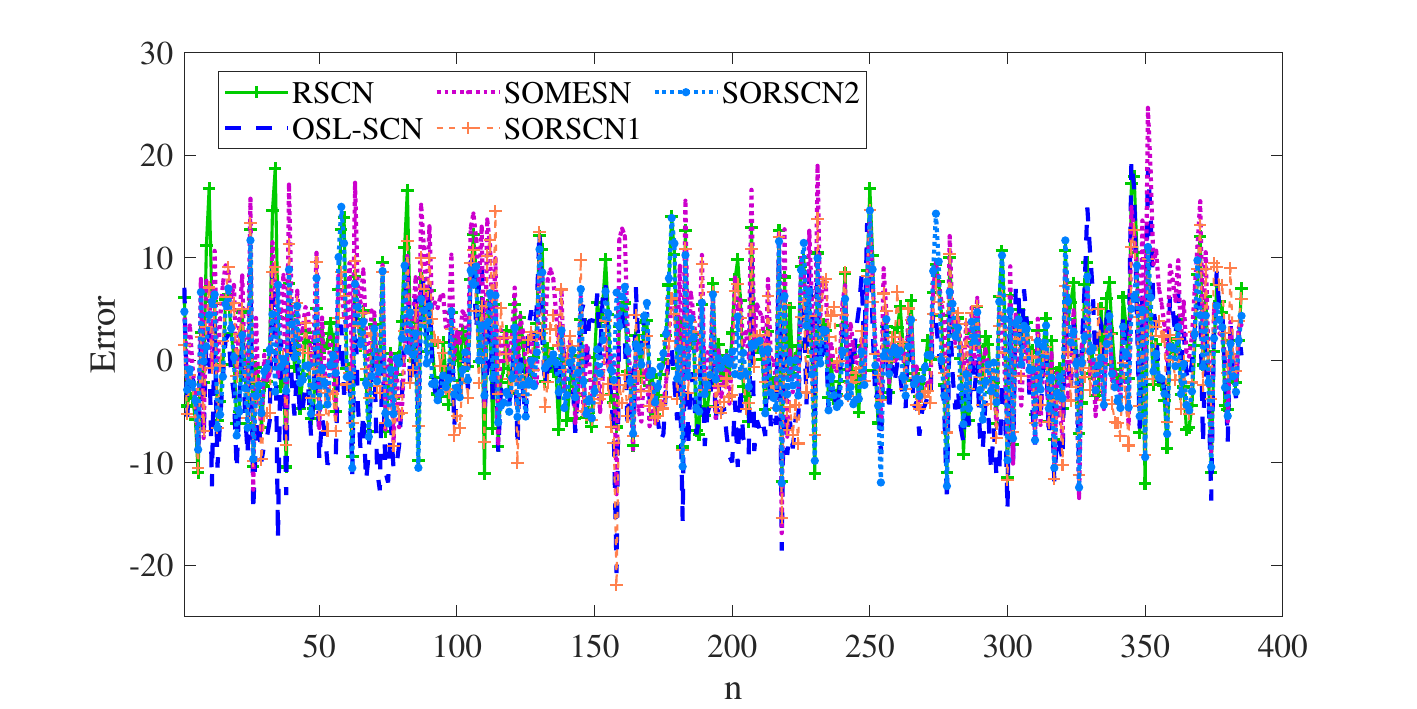}}
	\caption{Prediction fitting curves and error values of different models for the short-term power load forecasting.}
	\label{fig6}
 \vspace{-0.7cm}
\end{figure}

 Fig.~\ref{fig6} depicts the prediction curves and errors of different models for the short-term power load forecasting. The outputs generated by SORSCN2 are highly matched with the desired outputs and demonstrate the smallest prediction errors. These results validate the effectiveness of our proposed method in handling complex industrial processes, indicating their great potential for practical scenarios characterized by order uncertainty and nonstationary data streams.
 
\setlength{\abovecaptionskip}{5pt}

\begin{figure}[htbp]
	\begin{center}
		\includegraphics[width=8cm]{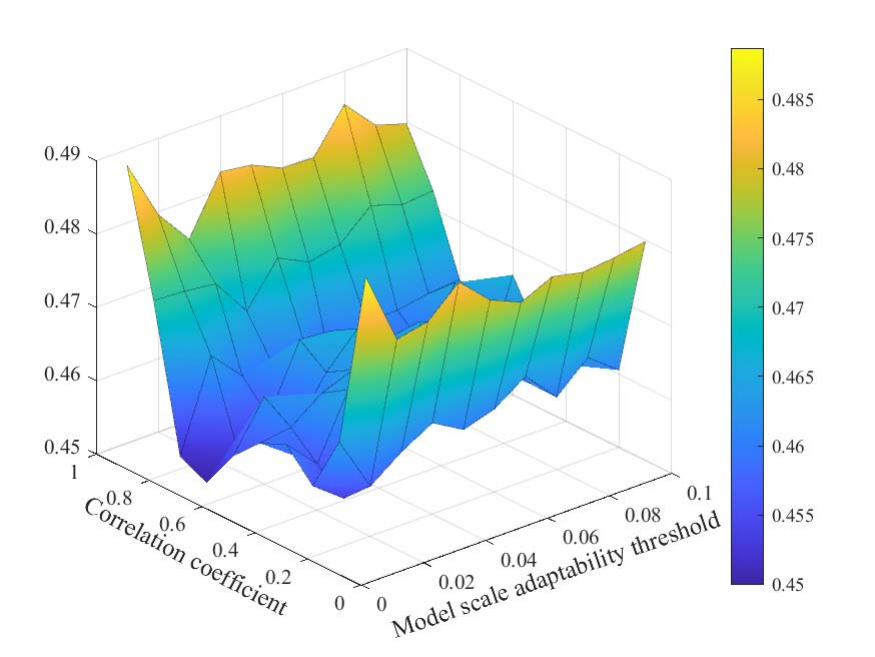}
		\caption{Testing NRMSE of SORSCN2 with different parameters (correlation coefficient $\alpha $  and the threshold of model scale adaptability $\gamma $ ) on the short-term power load forecasting task.}
		\label{fig611}
	\end{center}
 \vspace{-0.3cm}
\end{figure}

Fig.~\ref{fig611} displays the impact of correlation coefficient $\alpha $ and the threshold of model scale adaptability $\gamma $ on the short-term power load forecasting task. The proposed SORSCN2 has the best testing result when $\alpha=0.6$ and $\gamma=0.01$. Moreover, a correlation coefficient $\alpha$ that is either too large or too small can lead to poor prediction accuracy. These findings demonstrates that an appropriate $\alpha$ is important to obtain an accurate prediction model, verifying the effectiveness of introducing a correlation metric in our proposed method.

Table~\ref{tb2} presents a detailed performance comparison of various models on two industry cases. SORSCN2 exhibits superior performance in both validation and testing. Dynamically adjusting the network structure based on the supervisory mechanism enables SORSCNs to further improve their approximation performance when dealing with nonstationary data flows. These results indicate that the proposed SORSCNs offer notable benefits in modelling nonstationary dynamics in real-world industrial applications.

\begin{table}[h]
\vspace{-0.3cm}
\scriptsize
\caption{Performance comparison of different models on the two industry cases.} \label{tb2}
\centering
\begin{tabular}{cccccc}
\hline
Datasets                & Models & $N_{max}$  & $N$   & Validation NRMSE         & Testing NRMSE            \\ \hline
\multirow{6}{*}{Case 1} & ESN     & 300 & 213 & 0.0712±0.0092    & 0.0842±0.0114 \\
                        & RSCN    & 300 & 87  & 0.0633±0.0056    & 0.0679±0.0041 \\
                        & OSL-SCN   & 300 & 132 & 0.0628±0.0083    & 0.0656±0.0066 \\
                        & SOMESN  & 300 & 150 & 0.0416±0.0032    & 0.0439±0.0013 \\
                        & SORSCN1 & 300 & 100 & 0.0402±0.0037    & 0.0415±0.0012 \\
                        & SORSCN2 & 300 & 90  & \textbf{0.0371±0.0028}    & \textbf{0.0376±0.0009} \\ \hline
\multirow{6}{*}{Case 2} & ESN     & 200 & 103 & 0.3209±0.0283    & 0.6452±0.0717 \\
                        & RSCN    & 200 & 65  & 0.2559±0.0203    & 0.5107±0.0368 \\
                        & OSL-SCN   & 200 & 76  & 0.2639±0.0138    & 0.5239±0.0098 \\
                        & SOMESN  & 200 & 75  & 0.2536±0.0102    & 0.5473±0.0092 \\
                        & SORSCN1 & 200 & 65  & 0.2227±0.0061    & 0.4923±0.0079 \\
                        & SORSCN2 & 200 & 60  & \textbf{0.2203±0.0065}    & \textbf{0.4612±0.0045} \\ \hline
\end{tabular}
\vspace{-0.6cm}
\end{table}

\begin{figure}[htbp]
\centering
	\subfloat[Case 1 ($N_{max}=300$, the size of subreservoir and time window is set to 10 and 40.)]{\includegraphics[width=9cm]{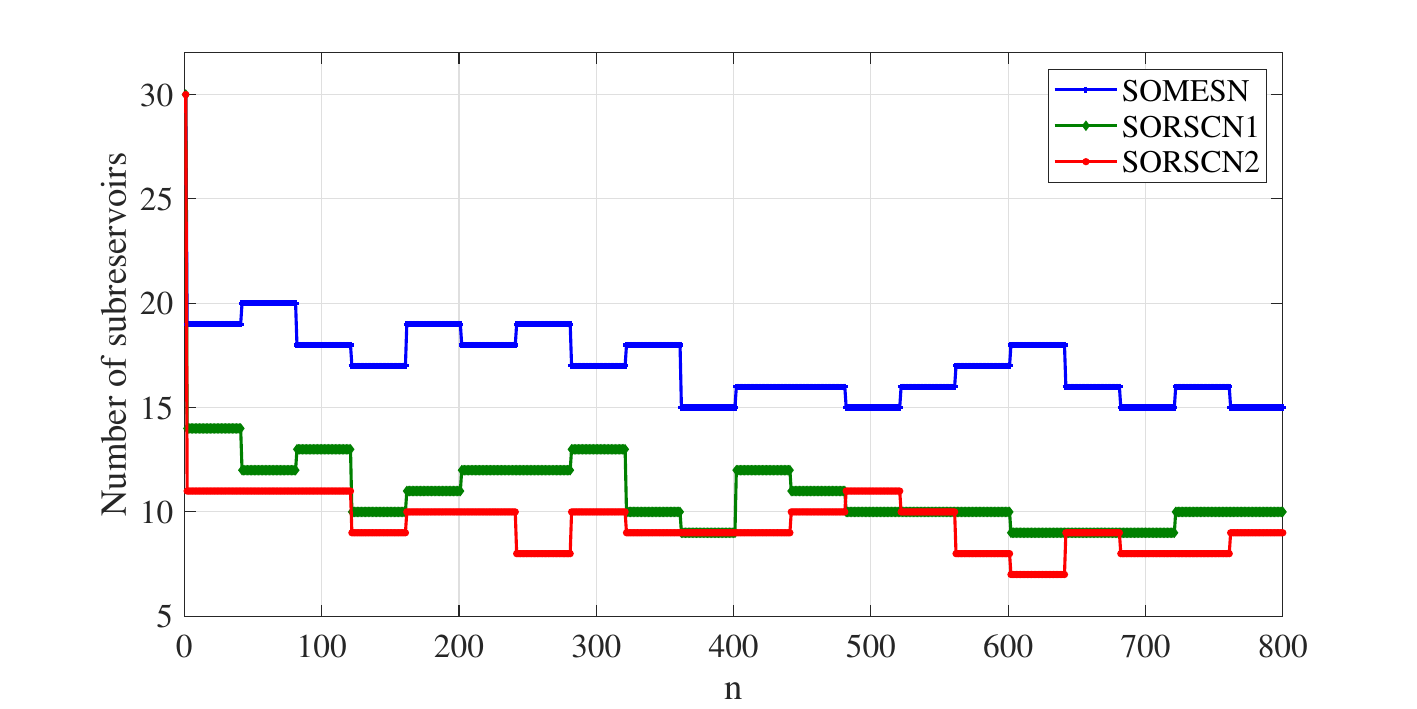}}\\ \vspace{-0.3cm}
	\subfloat[Case 2 ($N_{max}=200$, the size of subreservoir and time window is set to 10 and 20.)]{\includegraphics[width=9cm]{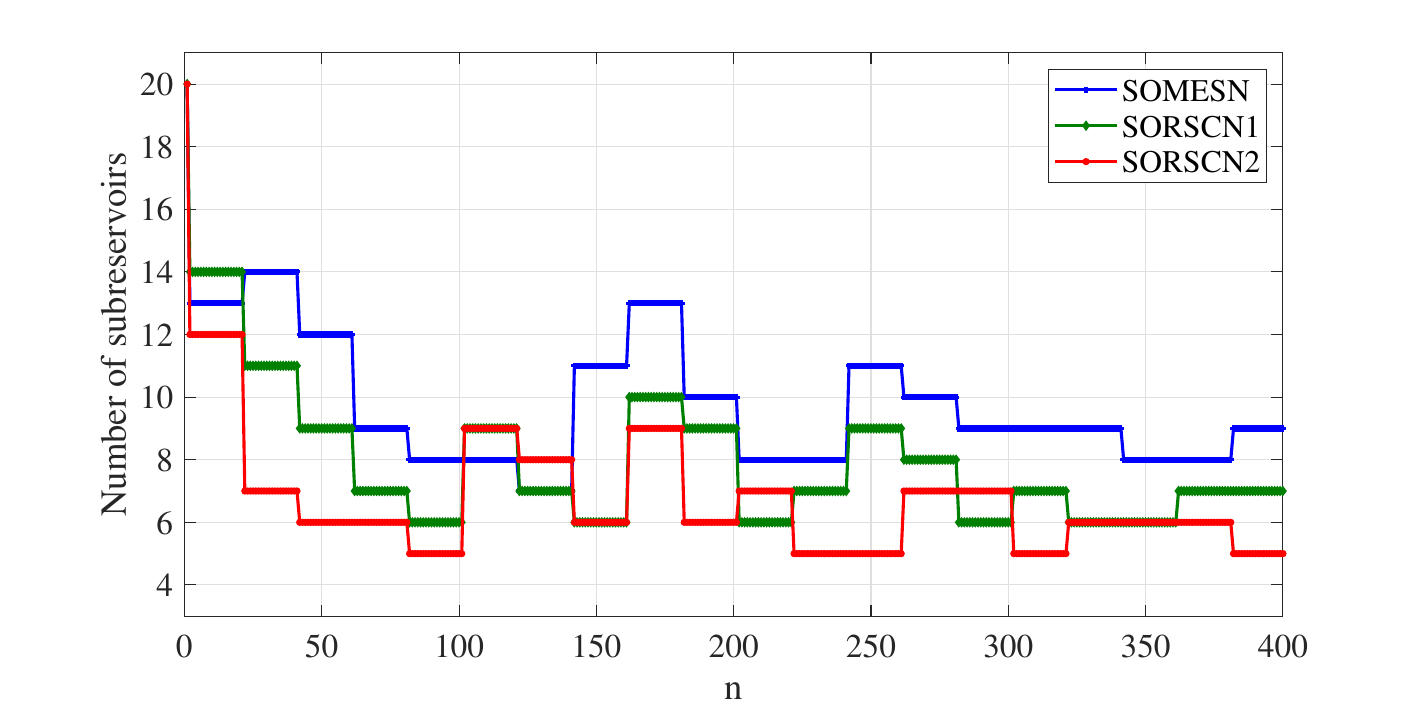}}
	\caption{Structure dynamic change diagram of the different self-organizing models on the validation datasets.}
		\label{fig7}
\vspace{-0.1cm}
\end{figure}

Fig.~\ref{fig7} demonstrates the structure self-learning process of the SOMESN and SORSCNs on the two industrial tasks. The self-organizing strategy allows for dynamic growing or pruning of subreservoirs based on real-time information, optimizing network structure and improving learning performance. Our proposed SORSCN2 utilizes the smallest number of subreservoirs and features a more compact reservoir topology. This highlights the potential of SORSCN2 in practical applications, especially in handling complex nonstationary industrial modelling tasks with flexibility and adaptability.\vspace{-0.25cm}

\subsection{Discussion}
Nonstationary dynamics are prevalent in industrial processes, necessitating that the built model be adaptive and flexible to the data flow. By incorporating self-organizing strategies, SORSCNs can automatically adjust their network structure based on real-time data, which includes modifying the number of subreservoirs and updating output weights. This dynamic learning mechanism allows the model to effectively capture the time-varying characteristics of the system while maintaining high prediction accuracy. Distinguished with SORSCN1, SORSCN2 introduces a correlation metric in the sensitivity analysis to eliminate redundant information, retain subreservoirs that are highly relevant to the target task, resulting a more compact and efficient network structure. Therefore, SORSCN2 is well-suited for resource-constrained environments, demonstrating faster response capability and improved predictive performance for the real-time data streams.
\vspace{-0.2cm}

\section{Conclusion}
Data streams generated from industrial processes are influenced by various complex factors, leading to dynamic time series that are highly nonlinear and nonstationary. This paper proposes a novel randomized learner model termed as self-organizing recurrent stochastic configuration network for problem resolving. SORSCNs adopt a hybrid strategy that integrates the online parameter self-learning and network structure self-adjustment. Based on the initially built model, the projection algorithm is used to online update output weights, and an improved sensitivity analysis with correlation metric is implemented to identify and remove subreservoirs that do not significantly contribute to the learning task. Finally, the RSC algorithm is employed to construct the network structure according to the newly arriving data streams. Experimental results demonstrate that our proposed SORSCNs outperform other models in terms of structure compactness and generalization, highlighting their great potential in modelling nonlinear dynamic systems with nonstationary dynamics.

Continual research is essential to develop more robust and effective growing or pruning algorithms that can improve the online adaptation and learning capabilities of the network. Additionally, further studies could investigate other advanced randomized learner models to enhance the logical reasoning ability and interpretability of the network.

\end{CJK}
\end{document}